\pdfoutput=1

\documentclass[11pt]{article}

\usepackage[preprint]{acl}

\usepackage{times}
\usepackage{latexsym}

\usepackage[T1]{fontenc}

\usepackage[utf8]{inputenc}

\usepackage{microtype}

\usepackage{inconsolata}

\usepackage{graphicx}
\usepackage{wrapfig}
\usepackage{makecell}
\usepackage{multirow}
\usepackage{multicol}
\usepackage{booktabs}
\usepackage{amssymb}

\usepackage{pythonhighlight}

\lstnewenvironment{PythonB}[1][]{\lstset{style=mypython, tabsize=4, frame=top, frame=bottom #1}}{}

\usepackage{tabularx} 
\usepackage{ragged2e} 
\usepackage{booktabs} 

\usepackage{threeparttable}

\usepackage{float}
\usepackage{stfloats}

\newcommand\blfootnote[1]{%
  \begingroup
  \renewcommand\thefootnote{}\footnote{#1}%
  \addtocounter{footnote}{-1}%
  \endgroup
}

\title{PCToolkit: A Unified Plug-and-Play Prompt Compression Toolkit of Large Language Models}

\author{Jinyi Li$^{1,3}$, Yihuai Lan$^{1}$, Lei Wang$^{2}$, Hao Wang$^{1*}$\\
   $^1$The Hong Kong University of Science and Technology (Guangzhou) \\
  $^2$Singapore Management University \\
  $^3$South China University of Technology \\
  \texttt{\{jinyili, haowang\}@hkust-gz.edu.cn} \\
  Open-source repository: \url{https://github.com/3DAgentWorld/Toolkit-for-Prompt-Compression}\\
  Supplementary video: \url{https://youtu.be/_KarBVRmpT0}
  }

\begin{document}
\maketitle
\blfootnote{$^*$Corresponding author.}
\begin{abstract}
Prompt compression is an innovative method for efficiently condensing input prompts while preserving essential information. To facilitate quick-start services, user-friendly interfaces, and compatibility with common datasets and metrics, we present the Prompt Compression Toolkit (PCToolkit). This toolkit is a unified plug-and-play solution for compressing prompts in Large Language Models (LLMs), featuring cutting-edge prompt compressors, diverse datasets, and metrics for comprehensive performance evaluation. PCToolkit boasts a modular design, allowing for easy integration of new datasets and metrics through portable and user-friendly interfaces. In this paper, we outline the key components and functionalities of PCToolkit.
We conducted evaluations of the compressors within PCToolkit across various natural language tasks, including reconstruction, summarization, mathematical problem-solving, question answering, few-shot learning, synthetic tasks, code completion, boolean expressions, multiple choice questions, and lies recognition.
\end{abstract}

\section{Introduction}

Given the performance limitations and computational overhead of Large Language Models (LLMs) \cite{Wang2024BeyondTL}, how to effectively apply LLMs to tasks involving lengthy textual inputs is a persistent challenge.
Various viable solutions have emerged to address this issue, encompassing techniques such as length extrapolation \cite{chen-etal-2021-simple, Shaw2018SelfAttentionWR}, attention approximation \cite{Winata2019LightweightAE, Wang2020LinformerSW}, attention-free transformers \cite{Gu2021EfficientlyML, Orvieto2023ResurrectingRN}, model compression \cite{lee2023owq, NEURIPS2023_44956951}, and hardware-aware transformers \cite{NEURIPS2022_67d57c32, NEURIPS2023_1bfd87d2}.

Prompt compression technology, a subset of length extrapolation methods, presents a strategic solution to tackle this challenge by condensing intricate textual inputs into succinct prompts that encapsulate crucial information. This approach enables LLMs to function more efficiently within resource constraints, enhancing their performance \cite{Wang2024BeyondTL}. Moreover, by reducing the reliance on extensive API calls, prompt compression not only improves the cost-effectiveness of leveraging LLMs but also streamlines the computational processes involved in language understanding tasks. When compared to alternative strategies, prompt compression offers intuitive and adaptable techniques for addressing diverse scenarios \cite{Naveed2023ACO, Zhao2023ASO, Wan2023EfficientLL}.

However, the deployment of prompt compression methods varies between different approaches. There is not yet a general toolkit that can invoke compressors of multiple types. Moreover, datasets and metrics are also essential for evaluating the performance of each compression method. Thus, with the aim of providing plug-and-play services, easy-customized interfaces and supporting common datasets and metrics, we propose Prompt Compression Toolkit (PCToolkit), a unified plug-and-play toolkit for Prompt Compression of LLMs, making accessible and portable prompt compression methods to a wider audience. Our plug-and-play design enables users to deploy and use the toolkit without any further model trainings. Meanwhile, users are also able to plug in their custom-trained models in PCToolkit.

\begin{figure*}
\centering
\includegraphics[width=150mm]{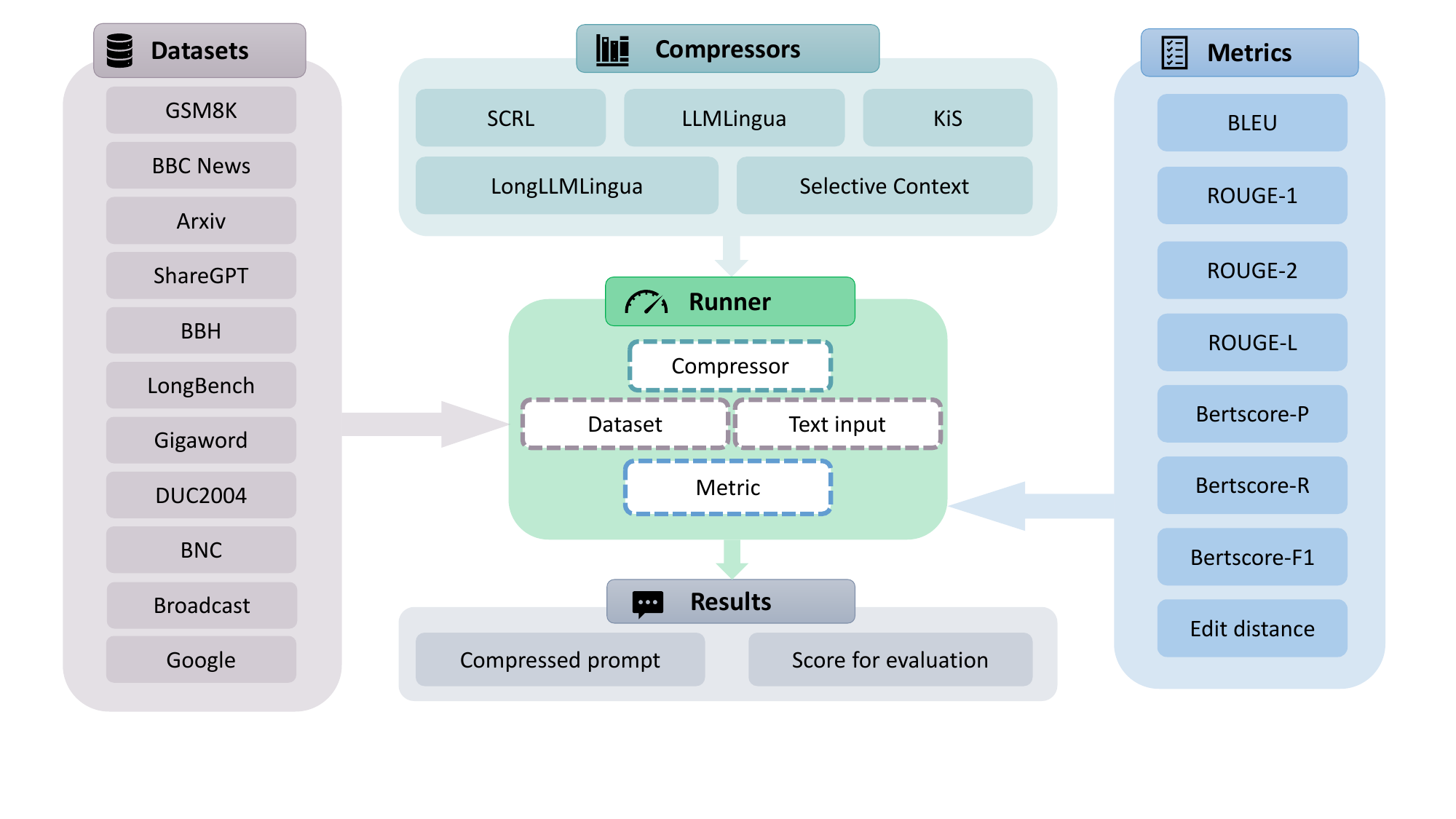}
\vspace{-30pt}
\caption{Architecture of PCToolkit. The \emph{compressors} module encompasses prompt compression methods that can be accessed through a unified interface with customizable parameters. The \emph{datasets} module includes 10 diverse datasets detailed in Table~\ref{datasets}. The \emph{metrics} module comprises four primary metrics utilized for evaluating the performance of various compressors. The \emph{runner} module offers a generalized interface for executing evaluations or simply retrieving the compressed prompt generated by the compressors.}
\label{fig1:env}
\end{figure*}

Specifically, Figure~\ref{fig1:env} illustrates the comprehensive architecture of PCToolkit. Key features of PCToolkit include:

(i) \textbf{State-of-the-art and reproducible methods.} Encompassing a wide array of mainstream compression techniques, PCToolkit offers a unified interface for various compression methods (compressors). Notably, PCToolkit incorporates a total of five distinct compressors, namely Selective Context \cite{Li2023CompressingCT}, LLMLingua \cite{Jiang2023LLMLinguaCP}, LongLLMLingua \cite{Jiang2023LongLLMLinguaAA}, SCRL \cite{ghalandari-etal-2022-efficient}, and KiS \cite{laban-etal-2021-keep}.

(ii) \textbf{User-friendly interfaces for new compressors, datasets, and metrics.} Facilitating portability and ease of adaptation to different environments, the interfaces within PCToolkit are designed to be easily customizable. This flexibility makes PCToolkit suitable for a wide range of environments and tasks.

(iii) \textbf{Modular design.} Featuring a modular structure that simplifies the transition between different methods, datasets, and metrics, PCToolkit is organized into four distinct modules: Compressor, Dataset, Metric and Runner module.

\section{Related Works}

Recent prompt-related toolkits have focused on prompt design intricacies and their influence on language model performance \cite{Amatriain2024PromptDA,Liu2021PretrainPA}. These studies emphasize the significance of tailored prompts in guiding language models for accurate information retrieval, offering valuable insights for prompt compression methodologies. Various toolkits exist for prompt engineering and optimization, such as Promptify, ChainForge, Promptotype, and OpenPrompt.

\textbf{Promptify.} It is a toolkit tailored for prompt engineering, addressing NLP challenges with LLMs and facilitating the generation of diverse NLP task prompts \cite{Promptify2022}.

\textbf{ChainForge.} This visual toolkit aids prompt engineering and enables on-demand hypothesis testing for text generation LLMs \cite{Arawjo2023ChainForgeAV}.

\textbf{Promptotype.} A platform for structured prompt engineering, facilitating the development, testing, and monitoring of customized LLM tasks\footnote{\url{https://www.promptotype.io/}}.

\textbf{OpenPrompt.} This toolkit supports prompt-learning with pre-trained language models (PLMs), offering efficiency, modularity, and extendibility. It allows the integration of different PLMs, task formats, and prompting modules in a unified framework \cite{ding-etal-2022-openprompt}.

Despite the availability of aforementioned toolkits, a toolkit specifically focusing on prompt compression remains absent. By amalgamating insights from existing works and incorporating state-of-the-art prompt compression techniques, our toolkit aims to equip researchers, developers, and practitioners with a versatile toolset for prompt compression. This enhancement seeks to improve the performance and affordability of large language models across diverse applications.

\begin{table*}
    \small
    \renewcommand\arraystretch{1.2}
    \centering
    \resizebox{2\columnwidth}{!}{
    \begin{tabular}{p{4cm}p{8cm}p{5cm}}
    \toprule
        \textbf{Tasks} & \textbf{Supported Compressors} & \textbf{Supported Datasets} \\ \hline
        Reconstruction & SC, LLMLingua, LongLLMLingua, SCRL, KiS & BBC, ShareGPT, Arxiv, GSM8K \\ \hline
        Mathematical promblems & SC, LLMLingua, LongLLMLingua, SCRL, KiS & GSM8K, BBH \\ \hline

        Boolean expressions & SC, LLMLingua, LongLLMLingua, SCRL, KiS & BBH \\ \hline
        Multiple choice & SC, LLMLingua, LongLLMLingua, SCRL, KiS & BBH \\ \hline
        Lies recognition & SC, LLMLingua, LongLLMLingua, SCRL, KiS & BBH \\ \hline
        
        \multirow{4}{*}{Summarization} & \multirow{2}{*}{SC, LLMLingua, LongLLMLingua, SCRL, KiS} & BBC, Arxiv. Gigaword, DUC2004, BNC, Broadcast, Google \\ 
        \cline{2-3}
        ~ & LLMLingua, LongLLMLingua & LongBench \\ \hline
        \multirow{2}{*}{Question and Answer} & SC, LLMLingua, LongLLMLingua, SCRL, KiS & BBH \\ 
        \cline{2-3}
        ~ & LLMLingua, LongLLMLingua & LongBench \\ \hline
        Few-shot learning & LLMLingua, LongLLMLingua & LongBench \\ \hline
        Synthetic tasks & LLMLingua, LongLLMLingua & LongBench \\ \hline
        Code completion & LLMLingua, LongLLMLingua & LongBench \\ 
        \bottomrule
    \end{tabular}
    }
\caption{\label{PCToolkitoverview}
An overview of PCToolkit, including different evaluation tasks, compressors and datasets. }   
\end{table*}


\section{Supported Compressors, Datasets and Metrics}

Table~\ref{PCToolkitoverview} presents an overview of the supported tasks, compressors, and datasets within PCToolkit. Each component are described in detail in \textbf{Section 4 Toolkit Design}. Evaluation of all compression methods across various datasets for different tasks is depicted in Table~\ref{ticks}, with results to be discussed in \textbf{Section 5 Evaluation}.


\subsection{Compressors}

PCToolkit integrates 5 state-of-the-art prompt compression methods in total: Selective Context \cite{Li2023CompressingCT}, LLMLingua \cite{Jiang2023LLMLinguaCP}, LongLLMLingua \cite{Jiang2023LongLLMLinguaAA}, SCRL \cite{ghalandari-etal-2022-efficient} and KiS \cite{laban-etal-2021-keep}. These compressors are plug-and-play implemented, therefore can be invoked directly.

\textbf{Selective Context.} Selective Context improves the context efficiency of LLMs in inference by removing redundant content measured by self-information  \cite{6773024}. 

\textbf{LLMLingua.} LLMLingua involves a budget controller to maintain semantic integrity under high compression ratios. LLMLingua compresses information within prompts by capitalizing on the compression-like characteristics of LLMs  \cite{Jiang2023LLMLinguaCP}.

\textbf{LongLLMLingua.} LongLLMLingua came into stage with an enhancement on dealing with the inherent challenge of the \emph{lost in the middle} issue  \cite{Liu2023LostIT}, which is a phenomenon that performance of LLM can degrade significantly when models must access relevant information in the middle of long contexts \cite{Jiang2023LongLLMLinguaAA}.

\textbf{SCRL.} SCRL is a reinforcement learning-based approach designed to remove or retain tokens according to the probabilities \cite{ghalandari-etal-2022-efficient}.

\textbf{KiS.} KiS is an approach of unsupervised text simplification, which learns to balance a reward across three properties: fluency, salience and simplicity \cite{laban-etal-2021-keep}.

\begin{table*}
\centering
\resizebox{1.5\columnwidth}{!}{
\begin{tabular}{lcp{7cm}}
\toprule
\textbf{Datasets} & \textbf{Supporting Compressors} & \textbf{Supporting Metrics} \\
\hline
BBH & All & Accuracy\\
Gigaword & All & ROUGE, Token-F1 \\
BNC & All & ROUGE, Token-F1 \\
DUC2004 & All & ROUGE, Token-F1 \\
Broadcast & All & ROUGE, Token-F1 \\
Google & All & ROUGE, Token-F1 \\
GSM8K & All & Accuracy, BLEU, ROUGE, BERTScore \\
BBC News & All & BLEU, ROUGE, BERTScore \\
Arxiv articles & All & BLEU, ROUGE, BERTScore \\
ShareGPT & All & BLEU, ROUGE, BERTScore \\
\multirow{2}{*}{LongBench} & \multirow{2}{*}{LLMLingua, LongLLMLingua} & Accuracy, BLEU, ROUGE, BERTScore, Edit-distance \\

\bottomrule
\end{tabular}
}
\caption{\label{datasets}
Datasets and corresponding compressors and metrics supported in PCToolkit.}
\end{table*}

\subsection{Datasets}

Table~\ref{datasets} shows all datasets supported in PCToolkit. 


\textbf{GSM8K.} GSM8K \cite{cobbe2021gsm8k} contains 8.5K high-quality linguistically diverse word problems in elementary school mathematics. Each item contains a problem and its solution.

\textbf{BBC News, Arxiv articles and ShareGPT.} \citet{Li2023CompressingCT} provided the three datasets. BBC News provides news articles from BBC, which is a typical context of human daily lives. Arxiv articles provides scientific articles that represents a formal context. ShareGPT contains contexts that is collected from human-AI conversations, which is a normal communication context.

\textbf{Big Bench Hard (BBH).} BBH \cite{Suzgun2022ChallengingBT} is a diverse evaluation suite that focuses on a suite of 23 challenging tasks from BIG-Bench that were found to be beyond the capabilities of current language models.

\textbf{LongBench.} LongBench \cite{Bai2023LongBenchAB} is the first benchmark for bilingual, multitask and comprehensive assessment of long context understanding capabilities of large language models. LongBench has six different task scenarios including single-document question \& answer, multi-document question \& answer, summarization, few-shot learning, synthetic tasks and code completion.

\textbf{Gigaword, BNC, DUC2004, Broadcast and Google.} \citet{ghalandari-etal-2022-efficient} provided the five datasets. While Gigaword \cite{rush-etal-2015-neural} and DUC2004 \cite{2004Introduction} contain abstractive ground truth summaries, the remaining three datasets \cite{filippova-altun-2013-overcoming, Clarke2008GlobalIF} have token-level extractive ground truth summaries. 


\subsection{Metrics}

PCToolkit provides different metrics, including BLEU, ROUGE, BERTScore, Edit distance and Accuracy. The first four metrics are used to compare the difference between two strings, while Accuracy judges the results provided by LLM with the ground truth answer.

\textbf{BLEU.} Proposed by \citet{papineni-etal-2002-bleu}, Bilingual Evaluation Understudy (BLEU) is a metric used to evaluate machine-translated text by comparing it to reference translations \cite{papineni-etal-2002-bleu, Li2023CompressingCT}.


\textbf{ROUGE.} Proposed by \citet{Lin2004ROUGEAP}, Recall-Oriented Understudy for Gisting Evaluation (ROUGE) is a set of metrics used for evaluating the quality of summaries produced by automatic summarization systems \cite{Lin2004ROUGEAP, Li2023CompressingCT, Bai2023LongBenchAB}.


\textbf{BERTScore.} Proposed by \citet{Zhang*2020BERTScore:}, BERTScore evaluates text similarity using contextual embeddings from BERT \cite{devlin2019bert}. It measures the similarity between reference and candidate sentences, providing a score between 0 and 1, where 1 indicates perfect semantic similarity \cite{Zhang*2020BERTScore:, Li2023CompressingCT}.

\textbf{Edit distance.} Edit distance (Levenshtein distance) is popularly used in code generation evaluation \cite{Svyatkovskiy2020IntelliCodeCC, 4160958}. Edit Distance is a metric used to quantify the difference between two sequences of strings \cite{Bai2023LongBenchAB}. 

\begin{table*}[h!t]
    \centering
    \resizebox{2\columnwidth}{!}{
    \begin{tabular}{p{3cm}ccclclllll}
    \toprule
        \multirow{2}{*}{Algorithms} & \multirow{2}{*}{Metrics} & \multicolumn{9}{c}{Datasets} \\ 
        
        \cline{3-11}
        
        ~ & ~ & GSM8K & BBC News & ShareGPT & Arxiv & Gigaword & DUC2004 & BNC & Broadcast & Google \\ \hline
        
        Selective Context & \multirow{4}{*}{BLEU} & 0.56 & \textbf{0.32} & \textbf{0.37}$_{(+0.12)}$ & \textbf{0.29} & 0.24 & 0.24 & 0.54 & 0.45 & \textbf{0.45} \\
        (Long)LLMLingua & ~ & \textbf{0.78} & 0.17 & 0.25$_{(-0.02)}$ & 0.12 & 0.20 & 0.21 & \textbf{0.69} & \textbf{0.81} & 0.41 \\
        SCRL & ~ & 0.34 & 0.02 & 0.27 & 0.05 & \textbf{0.26} & \textbf{0.25} & 0.55 & 0.45 & \textbf{0.45} \\
        KiS & ~ & 0.52 & 0.02 & 0.07 & 0.00 & 0.17 & 0.20 & 0.50 & 0.45 & 0.33 \\ \hline
        
        Selective Context & \multirow{4}{*}{ROUGE L} & 0.82 & \textbf{0.69} & \textbf{0.70}$_{(+0.33)}$ & \textbf{0.57} & 0.19 & 0.14 & 0.58 & 0.57 & 0.51 \\
        (Long)LLMLingua & ~ & \textbf{0.90} & 0.50 & 0.56$_{(+0.17)}$ & 0.42 & \textbf{0.21} & \textbf{0.17} & \textbf{0.82} & \textbf{0.90} & \textbf{0.57} \\
        SCRL & ~ & 0.53 & 0.27 & 0.58 & 0.24 & \textbf{0.21}$_{(-0.02)}$ & 0.13$_{(-0.09)}$ & 0.41$_{(-0.38)}$ & 0.41$_{(-0.41)}$ & 0.36$_{(-0.34)}$ \\
        KiS & ~ & 0.73 & 0.31 & 0.32 & 0.08 & 0.17 & 0.16 & 0.60 & 0.58 & 0.45 \\ \hline
        
        Selective Context & \multirow{4}{*}{Bertscore P} & 0.96 & \textbf{0.89} & \textbf{0.90} & \textbf{0.93} & 0.86 & 0.86 & 0.89 & 0.88 & 0.92 \\
        (Long)LLMLingua & ~ & \textbf{0.98} & 0.86 & 0.89 & 0.88 & \textbf{0.88} & \textbf{0.89} & \textbf{0.97} & \textbf{0.98} & \textbf{0.96} \\
        SCRL & ~ & 0.68 & 0.81 & 0.87 & 0.85 & 0.86 & 0.86 & 0.86 & 0.82 & 0.91 \\
        KiS & ~ & 0.95 & 0.83 & 0.81 & 0.80 & 0.87 & \textbf{0.89} & 0.93 & 0.93 & \textbf{0.95} \\ \hline
        
        Selective Context & \multirow{4}{*}{Bertscore R} & 0.97 & 0.91 & \textbf{0.92} & \textbf{0.92} & \textbf{0.83} & 0.84 & 0.90 & 0.90 & 0.87 \\
        (Long)LLMLingua & ~ & \textbf{0.98} & 0.89 & 0.91 & 0.90 & \textbf{0.83} & 0.85 & \textbf{0.94} & \textbf{0.96} & \textbf{0.89} \\
        SCRL & ~ & 0.70 & 0.88 & \textbf{0.92} & 0.86 & 0.82 & 0.82 & 0.85 & 0.83 & 0.85 \\
        KiS & ~ & 0.95 & \textbf{0.93} & 0.90 & 0.84 & 0.84 & \textbf{0.86} & 0.91 & 0.90 & \textbf{0.89} \\ \hline
        
        Selective Context & \multirow{4}{*}{Bertscore F1} & 0.97 & \textbf{0.90} & \textbf{0.91}$_{(+0.02)}$ & \textbf{0.92} & \textbf{0.85} & 0.85 & 0.89 & 0.89 & 0.89 \\
        (Long)LLMLingua & ~ & \textbf{0.98} & 0.88 & 0.90$_{(+0.05)}$ & 0.89 & \textbf{0.85} & \textbf{0.87} & \textbf{0.96} & \textbf{0.97} & \textbf{0.93} \\
        SCRL & ~ & 0.69 & 0.84 & 0.89 & 0.85 & 0.84 & 0.84 & 0.85$_{(+0.09)}$ & 0.82$_{(+0.03)}$ & 0.88$_{(+0.11)}$ \\
        KiS & ~ & 0.95 & 0.88 & 0.85 & 0.82 & \textbf{0.85} & \textbf{0.87} & 0.92 & 0.91 & 0.92 \\
        
        \bottomrule
        
    \end{tabular}
    }
    \caption{\label{bigger}
Performance measured for reconstruction and summarization tasks in PCToolkit. (Long)LLMLingua means we considered LLMLingua and LongLLMLingua together, since for small scale datas, these two compressors showed very slight differences. Numbers in parenthesis are the difference between the original results provided by former experiments and our results.
}
\end{table*}

\section{Toolkit Design}

\subsection{Modular Design}

As shown in Figure~\ref{fig1:env}, PCToolkit is designed with a modular architecture, consisting of Compressor, Dataset, Metrics and Runner. 

\textbf{Compressors.} \verb|pctoolkit.compressors| module in PCToolkit encompasses five state-of-the-art compression methods tailored for prompt optimization. All compressors can be invoked through a unified interface shown in \textbf{Section 4.2}.

\textbf{Datasets.} \verb|pctoolkit.datasets| module boasts a diverse collection of over ten datasets, each meticulously curated to cover a wide array of natural language tasks. As shown in Table~\ref{bigger}, from tasks like reconstruction, summarization, question answering, to more specialized domains such as code completion and lies recognition, the datasets in PCToolkit offer a comprehensive testing ground for assessing the efficacy of prompt compression techniques.

\textbf{Metrics.} \verb|pctoolkit.metrics| module plays a crucial role in quantifying the performance of the compression methods across different tasks. All metrics needed can be easily contained inside a list that tells the Runner which metrics are required measuring. 

\textbf{Runners.} \verb|pctoolkit.runners| module serves as the engine that drives the evaluation process, orchestrating the interaction between the compression methods, datasets, and evaluation metrics. Researchers and practitioners can seamlessly execute experiments, compare results, and analyze the performance of different compression techniques using the Runner component. This streamlined workflow ensures efficient experimentation and evaluation of prompt compression strategies within the toolkit.

By integrating these components, PCToolkit offers a comprehensive and user-friendly platform for prompt compression and evaluation, empowering researchers and practitioners to optimize prompts for enhanced model performance in natural language processing tasks.

\subsection{Unified Interface}

In PCToolkit, a unified interface for invoking prompt compression methods is provided. In the following example, we show how to simply invoke the compressing methods within few lines.

\begin{PythonB}
from pctoolkit.compressors import 
    PromptCompressor

compressor = PromptCompressor(
type='SCCompressor', device='cuda')

test_prompt = "test prompt"
ratio = 0.3
result = compressor.compressgo(
test_prompt, ratio)
print(result)

\end{PythonB}

Different parameters for compressors can be included inside \verb|compressgo|.

For simple compression task, one compressor is selected. Following the example given above, an original prompt is input to the compressor, and the compressor outputs the target compressed prompt. For datasets evaluation, one datasets and multiple metrics are selected, along with the compressor chosen, these three parts are deployed in Runner. The Runner will provide the evaluation results according to the metrics list, which includes all metrics expected. The following example shows how to modularistically use PCToolkit.

\begin{PythonB}
from pctoolkit.runners import run
from pctoolkit.datasets import 
    load_dataset
from pctoolkit.metrics import 
    load_metrics

compressor = PromptCompressor(
type='SCCompressor', device='cuda')
dataset_name = 'arxiv'
dataset = load_dataset(dataset_name)

run(compressor=compressor, 
dataset=dataset, 
metrics=load_metrics, ratio=0.1)
\end{PythonB}

Currently, the supporting datasets calls are implemented inside \verb|run|. Users can also following the format in \verb|run| to adapt their own datasets or metrics.



\section{Evaluation}

\textbf{Compression ratio.} Following \citet{Li2023CompressingCT}, we define the compression ratio to be the ratio of reduced context length comparing with the original context length. That is, ratio $ \rho = 1 - \frac{L_c}{L_O}$ where $ L_c $ represents the length of compressed context and $ L_o $ represents the length of original context. Compression ratio is an essential parameter that measures how much deletion is needed for a prompt.

\begin{table*}
    \centering
    \resizebox{1.75\columnwidth}{!}{
    \begin{tabular}{lcccccccc}
    \toprule
        \multirow{2}{*}{Compressors} & \multicolumn{8}{c}{LongBench\footnotemark[3]} \\ 

        \cline{2-9}
        
        ~ & SingleDoc & MultiDoc & Summ. & FewShot & Synth. & Code & AVG & Ratio \\ \hline
        
        LLMLingua & 0.30$_{(-0.01)}$  & 0.34$_{(-0.04)}$  & 0.22$_{(-0.04)}$  & 0.63$_{(-0.04)}$  & 0.11$_{(+0.03)}$  & 0.37$_{(-0.16)}$  & 0.33$_{(-0.04)}$  & 0.66  \\    
        LongLLMLingua & 0.41$_{(+0.01)}$  & 0.39$_{(-0.07)}$  & 0.22$_{(-0.05)}$  & 0.63$_{(-0.08)}$  & 0.75$_{(+0.22)}$  & 0.42$_{(-0.13)}$  & 0.47$_{(-0.02)}$  & 0.66  \\ \hline
        
        LLMLingua & 0.26$_{(+0.04)}$  & 0.36$_{(+0.04)}$  & 0.22$_{(-0.03)}$  & 0.60$_{(-0.01)}$  & 0.10$_{(0.00)}$  & 0.36$_{(-0.21)}$  & 0.32$_{(-0.03)}$  & 0.80  \\  
        LongLLMLingua & 0.38$_{(-0.01)}$  & 0.39$_{(-0.03)}$  & 0.22$_{(-0.05)}$  & 0.62$_{(-0.07)}$  & 0.60$_{(+0.04)}$  & 0.37$_{(-0.2)}$  & 0.43$_{(-0.05)}$  & 0.80  \\ 
        \bottomrule
    \end{tabular}
    }
    \caption{\label{longbench}
Performance measured in LongBench datasets. Numbers in parenthesis are the difference between the original results provided by former experiments and our results. We evaluated each task by the metric provided by LongBench.}
\end{table*}

\subsection{Short Context Tasks}



We conducted evaluations using different datasets as outlined in Table~\ref{PCToolkitoverview} and assessed them across various metrics. The results, presented in Table~\ref{bigger}, utilized metrics like BLEU, ROUGE, and BERTScore for testing tasks that do not have a definitive answer, such as reconstruction and summarization. Following the methodologies of \citet{Li2023CompressingCT} and \citet{Jiang2023LLMLinguaCP}, GPT-3.5-Turbo was employed as a frozen LLM for reconstruction tasks. It received compressed prompts from the compressors and generated reconstructed prompts, which were then compared with the compressed ones. For summarization tasks, the frozen LLM provided a pair of summaries, one from the original context and the other from the compressed context. These pairs of summaries were evaluated using the specified metrics. In our experiment, datasets like GSM8K, BBC News, and ShareGPT were designated for reconstruction tasks, while the rest were assigned to summarization tasks.

For tasks with precise answers, such as mathematical problems, metrics like accuracy and edit distance are commonly used. As shown in Table~\ref{bbhGSM}, we tested all compression methods across various task types. The GSM8K dataset includes mathematical problems, while BBH encompasses a diverse range of tasks. For instance, the Boolean Expression task requires the LLM to provide answers to specific logical expressions; the Movie Recommendation task tasks the LLM with selecting the most suitable movie from a list based on a given description; and the Web of Lies task involves the LLM determining if a particular character is lying.

\begin{table}[h!t]
    \centering
    \resizebox{0.8\columnwidth}{!}{
    \begin{tabular}{lp{3cm}c}
    \toprule
        \textbf{Compressors} & \textbf{Dataset} & \textbf{Accuracy} \\ \hline

        Baseline & \multirow{5}{3cm}{BBH Boolean Expression} & 0.51 \\ 
        \cline{3-3} 
        Selective Context & ~ & \textbf{0.54}$_{(+0.03)}$ \\ 
        (Long)LLMLingua & ~ & \textbf{0.54}$_{(+0.03)}$ \\ 
        SCRL & ~ & \textbf{0.54}$_{(+0.03)}$ \\ 
        KiS & ~ & \textbf{0.54}$_{(+0.03)}$ \\ 
        \hline

        Baseline & \multirow{5}{3cm}{BBH Movie Recommendation} & 0.33 \\ 
        \cline{3-3} 
        Selective Context & ~ & 0.63$_{(+0.30)}$ \\ 
        (Long)LLMLingua & ~ & \textbf{0.67}$_{(+0.34)}$ \\
        SCRL & ~ & 0.59$_{(+0.26)}$ \\ 
        KiS & ~ & 0.48$_{(+0.15)}$ \\ 
        \hline

        Baseline & \multirow{5}{3cm}{BBH Web of Lies} & 0.89 \\ 
        \cline{3-3} 
        Selective Context & ~ & 0.39$_{(-0.50)}$ \\ 
        (Long)LLMLingua & ~ & \textbf{0.62}$_{(-0.27)}$ \\ 
        SCRL & ~ & 0.31$_{(-0.58)}$ \\ 
        KiS & ~ & 0.41$_{(-0.48)}$ \\ 
        \hline

        Baseline & \multirow{5}{3cm}{GSM8K} & 0.29 \\
        \cline{3-3} 
        Selective Context & ~ & 0.09$_{(-0.20)}$ \\ 
        (Long)LLMLingua & ~ & \textbf{0.25}$_{(-0.04)}$ \\ 
        SCRL & ~ & 0.05$_{(-0.24)}$ \\ 
        KiS & ~ & 0.13$_{(-0.16)}$ \\ 
        \bottomrule

    \end{tabular}
    }
    \caption{\label{bbhGSM}
Performance measured in BBH \& GSM8K datasets. Our baseline is the performance without using any compression methods. Numbers in parenthesis are the difference between the baseline results and each results with different compressors.}
\end{table}

\subsection{Long Context Tasks}

For datasets that contains longer contexts, we evaluate LLMLingua and LongLLMLingua on them. With specified questions, LongLLMLingua performed much better than LLMLingua. Results are shown in Table~\ref{longbench}. The evaluation settings are different from original ones, as the authors of LLMLingua mentioned on GitHub\footnote{ \url{https://github.com/microsoft/LLMLingua/blob/main/Transparency_FAQ.md}}, they used the completion mode of GPT-3.5-turbo, which is recently disabled by OpenAI. Thus, we used the chat mode instead, which caused a little deviation from the original results.

\section{Conclusion and Future work}



In conclusion, we introduced PCToolkit, an open-source project designed for prompt compression and evaluation.  This toolkit offers researchers and practitioners a user-friendly and comprehensive resource, featuring five cutting-edge compression methods and over ten diverse datasets encompassing a wide range of natural language tasks.  Through rigorous evaluations across various tasks such as reconstruction, summarization, mathematical problem-solving, question answering, few-shot learning, and more, we demonstrated the effectiveness and versatility of the compression techniques integrated into PCToolkit.

Our future endeavors focus on expanding PCToolkit with more compression methods, datasets, and evaluation metrics to further enhance its capabilities for prompt compression and model optimization in natural language processing.

\footnotetext[3]{\url{https://github.com/THUDM/LongBench}}



\section{Broader Impact}

The findings and methodologies presented in this study have broader implications for the field of natural language processing (NLP) and the development of language models.  By exploring and evaluating various compression techniques within the PCToolkit, we contribute to the ongoing efforts to enhance the efficiency and performance of large-scale language models.  The insights gained from this research can potentially inform the design of more streamlined and effective compression methods, paving the way for advancements in NLP applications across diverse domains.

Furthermore, the development of optimized compression methods could lead to more sustainable and eco-friendly practices in AI research and deployment.  By reducing the computational resources required for training and inference, we may contribute to a more energy-efficient and cost-effective utilization of AI technologies.

\section{Limitations}

Despite the advancements made in this study, there are inherent limitations that should be acknowledged.  One notable limitation is that the PCToolkit, while effective in compressing prompts and enhancing model performance, may still face challenges in handling toxic or harmful content present in NLP datasets.  The toolkit's current capabilities may not extend to effectively filtering out such content, highlighting the ongoing need for robust ethical guidelines and content moderation strategies in NLP research.

Additionally, the generalizability of the compression techniques evaluated in this study may be limited to specific task domains or dataset characteristics.  Further research is needed to explore the scalability and adaptability of these methods across a wider range of tasks and datasets to fully assess their utility and effectiveness in diverse applications.

Overall, while the PCToolkit offers valuable tools for prompt compression and model optimization, researchers and practitioners are encouraged to remain vigilant about the broader impacts and limitations associated with the use of such technologies in NLP research and development.

\bibliography{custom}

\begin{thebibliography}{39}
\expandafter\ifx\csname natexlab\endcsname\relax\def\natexlab#1{#1}\fi

\bibitem[{Amatriain(2024)}]{Amatriain2024PromptDA}
Xavier Amatriain. 2024.
\newblock \href {https://api.semanticscholar.org/CorpusID:267301483} {Prompt design and engineering: Introduction and advanced methods}.
\newblock \emph{ArXiv}, abs/2401.14423.

\bibitem[{Arawjo et~al.(2023)Arawjo, Swoopes, Vaithilingam, Wattenberg, and Glassman}]{Arawjo2023ChainForgeAV}
Ian Arawjo, Chelse Swoopes, Priyan Vaithilingam, Martin Wattenberg, and Elena~L. Glassman. 2023.
\newblock \href {https://api.semanticscholar.org/CorpusID:262044762} {Chainforge: A visual toolkit for prompt engineering and llm hypothesis testing}.
\newblock \emph{ArXiv}, abs/2309.09128.

\bibitem[{Bai et~al.(2023)Bai, Lv, Zhang, Lyu, Tang, Huang, Du, Liu, Zeng, Hou, Dong, Tang, and Li}]{Bai2023LongBenchAB}
Yushi Bai, Xin Lv, Jiajie Zhang, Hong Lyu, Jiankai Tang, Zhidian Huang, Zhengxiao Du, Xiao Liu, Aohan Zeng, Lei Hou, Yuxiao Dong, Jie Tang, and Juanzi Li. 2023.
\newblock \href {https://api.semanticscholar.org/CorpusID:261245264} {Longbench: A bilingual, multitask benchmark for long context understanding}.
\newblock \emph{ArXiv}, abs/2308.14508.

\bibitem[{Chen et~al.(2021)Chen, Tsai, Bhojanapalli, Chung, Chang, and Ferng}]{chen-etal-2021-simple}
Pu-Chin Chen, Henry Tsai, Srinadh Bhojanapalli, Hyung~Won Chung, Yin-Wen Chang, and Chun-Sung Ferng. 2021.
\newblock \href {https://doi.org/10.18653/v1/2021.emnlp-main.236} {A simple and effective positional encoding for transformers}.
\newblock In \emph{Proceedings of the 2021 Conference on Empirical Methods in Natural Language Processing}, pages 2974--2988, Online and Punta Cana, Dominican Republic. Association for Computational Linguistics.

\bibitem[{Clarke and Lapata(2008)}]{Clarke2008GlobalIF}
James Clarke and Mirella Lapata. 2008.
\newblock \href {https://api.semanticscholar.org/CorpusID:3004447} {Global inference for sentence compression : an integer linear programming approach}.
\newblock \emph{J. Artif. Intell. Res.}, 31:399--429.

\bibitem[{Cobbe et~al.(2021)Cobbe, Kosaraju, Bavarian, Chen, Jun, Kaiser, Plappert, Tworek, Hilton, Nakano, Hesse, and Schulman}]{cobbe2021gsm8k}
Karl Cobbe, Vineet Kosaraju, Mohammad Bavarian, Mark Chen, Heewoo Jun, Lukasz Kaiser, Matthias Plappert, Jerry Tworek, Jacob Hilton, Reiichiro Nakano, Christopher Hesse, and John Schulman. 2021.
\newblock Training verifiers to solve math word problems.
\newblock \emph{arXiv preprint arXiv:2110.14168}.

\bibitem[{Dao et~al.(2022)Dao, Fu, Ermon, Rudra, and R\'{e}}]{NEURIPS2022_67d57c32}
Tri Dao, Dan Fu, Stefano Ermon, Atri Rudra, and Christopher R\'{e}. 2022.
\newblock \href {https://proceedings.neurips.cc/paper_files/paper/2022/file/67d57c32e20fd0a7a302cb81d36e40d5-Paper-Conference.pdf} {Flashattention: Fast and memory-efficient exact attention with io-awareness}.
\newblock In \emph{Advances in Neural Information Processing Systems}, volume~35, pages 16344--16359. Curran Associates, Inc.

\bibitem[{Devlin et~al.(2019)Devlin, Chang, Lee, and Toutanova}]{devlin2019bert}
Jacob Devlin, Ming-Wei Chang, Kenton Lee, and Kristina Toutanova. 2019.
\newblock \href {http://arxiv.org/abs/1810.04805} {Bert: Pre-training of deep bidirectional transformers for language understanding}.

\bibitem[{Ding et~al.(2022)Ding, Hu, Zhao, Chen, Liu, Zheng, and Sun}]{ding-etal-2022-openprompt}
Ning Ding, Shengding Hu, Weilin Zhao, Yulin Chen, Zhiyuan Liu, Haitao Zheng, and Maosong Sun. 2022.
\newblock \href {https://doi.org/10.18653/v1/2022.acl-demo.10} {{O}pen{P}rompt: An open-source framework for prompt-learning}.
\newblock In \emph{Proceedings of the 60th Annual Meeting of the Association for Computational Linguistics: System Demonstrations}, pages 105--113, Dublin, Ireland. Association for Computational Linguistics.

\bibitem[{Filippova and Altun(2013)}]{filippova-altun-2013-overcoming}
Katja Filippova and Yasemin Altun. 2013.
\newblock \href {https://aclanthology.org/D13-1155} {Overcoming the lack of parallel data in sentence compression}.
\newblock In \emph{Proceedings of the 2013 Conference on Empirical Methods in Natural Language Processing}, pages 1481--1491, Seattle, Washington, USA. Association for Computational Linguistics.

\bibitem[{Ghalandari et~al.(2022)Ghalandari, Hokamp, and Ifrim}]{ghalandari-etal-2022-efficient}
Demian Ghalandari, Chris Hokamp, and Georgiana Ifrim. 2022.
\newblock \href {https://doi.org/10.18653/v1/2022.acl-long.90} {Efficient unsupervised sentence compression by fine-tuning transformers with reinforcement learning}.
\newblock In \emph{Proceedings of the 60th Annual Meeting of the Association for Computational Linguistics (Volume 1: Long Papers)}, pages 1267--1280, Dublin, Ireland. Association for Computational Linguistics.

\bibitem[{Gu et~al.(2021)Gu, Goel, and R'e}]{Gu2021EfficientlyML}
Albert Gu, Karan Goel, and Christopher R'e. 2021.
\newblock \href {https://api.semanticscholar.org/CorpusID:240354066} {Efficiently modeling long sequences with structured state spaces}.
\newblock \emph{ArXiv}, abs/2111.00396.

\bibitem[{Jiang et~al.(2023{\natexlab{a}})Jiang, Wu, Lin, Yang, and Qiu}]{Jiang2023LLMLinguaCP}
Huiqiang Jiang, Qianhui Wu, Chin-Yew Lin, Yuqing Yang, and Lili Qiu. 2023{\natexlab{a}}.
\newblock \href {https://api.semanticscholar.org/CorpusID:263830701} {Llmlingua: Compressing prompts for accelerated inference of large language models}.
\newblock In \emph{Conference on Empirical Methods in Natural Language Processing}.

\bibitem[{Jiang et~al.(2023{\natexlab{b}})Jiang, Wu, Luo, Li, Lin, Yang, and Qiu}]{Jiang2023LongLLMLinguaAA}
Huiqiang Jiang, Qianhui Wu, Xufang Luo, Dongsheng Li, Chin-Yew Lin, Yuqing Yang, and Lili Qiu. 2023{\natexlab{b}}.
\newblock \href {https://api.semanticscholar.org/CorpusID:263830692} {Longllmlingua: Accelerating and enhancing llms in long context scenarios via prompt compression}.
\newblock \emph{ArXiv}, abs/2310.06839.

\bibitem[{Laban et~al.(2021)Laban, Schnabel, Bennett, and Hearst}]{laban-etal-2021-keep}
Philippe Laban, Tobias Schnabel, Paul Bennett, and Marti~A. Hearst. 2021.
\newblock \href {https://doi.org/10.18653/v1/2021.acl-long.498} {Keep it simple: Unsupervised simplification of multi-paragraph text}.
\newblock In \emph{Proceedings of the 59th Annual Meeting of the Association for Computational Linguistics and the 11th International Joint Conference on Natural Language Processing (Volume 1: Long Papers)}, pages 6365--6378, Online. Association for Computational Linguistics.

\bibitem[{Lee et~al.(2023)Lee, Jin, Kim, Kim, and Park}]{lee2023owq}
Changhun Lee, Jungyu Jin, Taesu Kim, Hyungjun Kim, and Eunhyeok Park. 2023.
\newblock Owq: Lessons learned from activation outliers for weight quantization in large language models.
\newblock \emph{arXiv preprint arXiv:2306.02272}.

\bibitem[{Li et~al.(2023)Li, Dong, Lin, and Guerin}]{Li2023CompressingCT}
Yucheng Li, Bo~Dong, Chenghua Lin, and Frank Guerin. 2023.
\newblock \href {https://api.semanticscholar.org/CorpusID:263830231} {Compressing context to enhance inference efficiency of large language models}.
\newblock In \emph{Conference on Empirical Methods in Natural Language Processing}.

\bibitem[{Lin(2004)}]{Lin2004ROUGEAP}
Chin-Yew Lin. 2004.
\newblock \href {https://api.semanticscholar.org/CorpusID:964287} {Rouge: A package for automatic evaluation of summaries}.
\newblock In \emph{Annual Meeting of the Association for Computational Linguistics}.

\bibitem[{Liu and Abbeel(2023)}]{NEURIPS2023_1bfd87d2}
Hao Liu and Pieter Abbeel. 2023.
\newblock \href {https://proceedings.neurips.cc/paper_files/paper/2023/file/1bfd87d2d92f0556819467dc08034f76-Paper-Conference.pdf} {Blockwise parallel transformers for large context models}.
\newblock In \emph{Advances in Neural Information Processing Systems}, volume~36, pages 8828--8844. Curran Associates, Inc.

\bibitem[{Liu et~al.(2023)Liu, Lin, Hewitt, Paranjape, Bevilacqua, Petroni, and Liang}]{Liu2023LostIT}
Nelson~F. Liu, Kevin Lin, John Hewitt, Ashwin Paranjape, Michele Bevilacqua, Fabio Petroni, and Percy Liang. 2023.
\newblock \href {https://api.semanticscholar.org/CorpusID:259360665} {Lost in the middle: How language models use long contexts}.
\newblock \emph{Transactions of the Association for Computational Linguistics}, 12:157--173.

\bibitem[{Liu et~al.(2021)Liu, Yuan, Fu, Jiang, Hayashi, and Neubig}]{Liu2021PretrainPA}
Pengfei Liu, Weizhe Yuan, Jinlan Fu, Zhengbao Jiang, Hiroaki Hayashi, and Graham Neubig. 2021.
\newblock \href {https://api.semanticscholar.org/CorpusID:236493269} {Pre-train, prompt, and predict: A systematic survey of prompting methods in natural language processing}.
\newblock \emph{ACM Computing Surveys}, 55:1 -- 35.

\bibitem[{Ma et~al.(2023)Ma, Fang, and Wang}]{NEURIPS2023_44956951}
Xinyin Ma, Gongfan Fang, and Xinchao Wang. 2023.
\newblock \href {https://proceedings.neurips.cc/paper_files/paper/2023/file/44956951349095f74492a5471128a7e0-Paper-Conference.pdf} {Llm-pruner: On the structural pruning of large language models}.
\newblock In \emph{Advances in Neural Information Processing Systems}, volume~36, pages 21702--21720. Curran Associates, Inc.

\bibitem[{Naveed et~al.(2023)Naveed, Khan, Qiu, Saqib, Anwar, Usman, Barnes, and Mian}]{Naveed2023ACO}
Humza Naveed, Asad~Ullah Khan, Shi Qiu, Muhammad Saqib, Saeed Anwar, Muhammad Usman, Nick Barnes, and Ajmal~S. Mian. 2023.
\newblock \href {https://api.semanticscholar.org/CorpusID:259847443} {A comprehensive overview of large language models}.
\newblock \emph{ArXiv}, abs/2307.06435.

\bibitem[{Orvieto et~al.(2023)Orvieto, Smith, Gu, Fernando, Gulcehre, Pascanu, and De}]{Orvieto2023ResurrectingRN}
Antonio Orvieto, Samuel~L. Smith, Albert Gu, Anushan Fernando, Caglar Gulcehre, Razvan Pascanu, and Soham De. 2023.
\newblock \href {https://api.semanticscholar.org/CorpusID:257496654} {Resurrecting recurrent neural networks for long sequences}.
\newblock \emph{ArXiv}, abs/2303.06349.

\bibitem[{Over and Yen.(2004)}]{2004Introduction}
P~Over and J~Yen. 2004.
\newblock Introduction to duc 2004: an intrinsic evaluation of generic news text summarization systems.
\newblock In \emph{Document Understanding Conference}.

\bibitem[{Pal(2022)}]{Promptify2022}
Ankit Pal. 2022.
\newblock Promptify: Structured output from llms.
\newblock \url{https://github.com/promptslab/Promptify}.
\newblock Prompt-Engineering components for NLP tasks in Python.

\bibitem[{Papineni et~al.(2002)Papineni, Roukos, Ward, and Zhu}]{papineni-etal-2002-bleu}
Kishore Papineni, Salim Roukos, Todd Ward, and Wei-Jing Zhu. 2002.
\newblock \href {https://doi.org/10.3115/1073083.1073135} {{B}leu: a method for automatic evaluation of machine translation}.
\newblock In \emph{Proceedings of the 40th Annual Meeting of the Association for Computational Linguistics}, pages 311--318, Philadelphia, Pennsylvania, USA. Association for Computational Linguistics.

\bibitem[{Rush et~al.(2015)Rush, Chopra, and Weston}]{rush-etal-2015-neural}
Alexander~M. Rush, Sumit Chopra, and Jason Weston. 2015.
\newblock \href {https://doi.org/10.18653/v1/D15-1044} {A neural attention model for abstractive sentence summarization}.
\newblock In \emph{Proceedings of the 2015 Conference on Empirical Methods in Natural Language Processing}, pages 379--389, Lisbon, Portugal. Association for Computational Linguistics.

\bibitem[{Shannon(1948)}]{6773024}
C.~E. Shannon. 1948.
\newblock \href {https://doi.org/10.1002/j.1538-7305.1948.tb01338.x} {A mathematical theory of communication}.
\newblock \emph{The Bell System Technical Journal}, 27(3):379--423.

\bibitem[{Shaw et~al.(2018)Shaw, Uszkoreit, and Vaswani}]{Shaw2018SelfAttentionWR}
Peter Shaw, Jakob Uszkoreit, and Ashish Vaswani. 2018.
\newblock \href {https://api.semanticscholar.org/CorpusID:3725815} {Self-attention with relative position representations}.
\newblock In \emph{North American Chapter of the Association for Computational Linguistics}.

\bibitem[{Suzgun et~al.(2022)Suzgun, Scales, Scharli, Gehrmann, Tay, Chung, Chowdhery, Le, hsin Chi, Zhou, and Wei}]{Suzgun2022ChallengingBT}
Mirac Suzgun, Nathan Scales, Nathanael Scharli, Sebastian Gehrmann, Yi~Tay, Hyung~Won Chung, Aakanksha Chowdhery, Quoc~V. Le, Ed~Huai hsin Chi, Denny Zhou, and Jason Wei. 2022.
\newblock \href {https://api.semanticscholar.org/CorpusID:252917648} {Challenging big-bench tasks and whether chain-of-thought can solve them}.
\newblock In \emph{Annual Meeting of the Association for Computational Linguistics}.

\bibitem[{Svyatkovskiy et~al.(2020)Svyatkovskiy, Deng, Fu, and Sundaresan}]{Svyatkovskiy2020IntelliCodeCC}
Alexey Svyatkovskiy, Shao~Kun Deng, Shengyu Fu, and Neel Sundaresan. 2020.
\newblock \href {https://api.semanticscholar.org/CorpusID:218673683} {Intellicode compose: code generation using transformer}.
\newblock \emph{Proceedings of the 28th ACM Joint Meeting on European Software Engineering Conference and Symposium on the Foundations of Software Engineering}.

\bibitem[{Wan et~al.(2023)Wan, Wang, Liu, Alam, Zheng, Liu, Qu, Yan, Zhu, Zhang, Chowdhury, and Zhang}]{Wan2023EfficientLL}
Zhongwei Wan, Xin Wang, Che Liu, Samiul Alam, Yu~Zheng, Jiachen Liu, Zhongnan Qu, Shen Yan, Yi~Zhu, Quanlu Zhang, Mosharaf Chowdhury, and Mi~Zhang. 2023.
\newblock \href {https://api.semanticscholar.org/CorpusID:266044196} {Efficient large language models: A survey}.
\newblock \emph{ArXiv}, abs/2312.03863.

\bibitem[{Wang et~al.(2020)Wang, Li, Khabsa, Fang, and Ma}]{Wang2020LinformerSW}
Sinong Wang, Belinda~Z. Li, Madian Khabsa, Han Fang, and Hao Ma. 2020.
\newblock \href {https://api.semanticscholar.org/CorpusID:219530577} {Linformer: Self-attention with linear complexity}.
\newblock \emph{ArXiv}, abs/2006.04768.

\bibitem[{Wang et~al.(2024)Wang, Salmani, Omidi, Ren, Rezagholizadeh, and Eshaghi}]{Wang2024BeyondTL}
Xindi Wang, Mahsa Salmani, Parsa Omidi, Xiangyu Ren, Mehdi Rezagholizadeh, and Armaghan Eshaghi. 2024.
\newblock \href {https://api.semanticscholar.org/CorpusID:267412232} {Beyond the limits: A survey of techniques to extend the context length in large language models}.
\newblock \emph{ArXiv}, abs/2402.02244.

\bibitem[{Winata et~al.(2019)Winata, Cahyawijaya, Lin, Liu, and Fung}]{Winata2019LightweightAE}
Genta~Indra Winata, Samuel Cahyawijaya, Zhaojiang Lin, Zihan Liu, and Pascale Fung. 2019.
\newblock \href {https://api.semanticscholar.org/CorpusID:204960988} {Lightweight and efficient end-to-end speech recognition using low-rank transformer}.
\newblock \emph{ICASSP 2020 - 2020 IEEE International Conference on Acoustics, Speech and Signal Processing (ICASSP)}, pages 6144--6148.

\bibitem[{Yujian and Bo(2007)}]{4160958}
Li~Yujian and Liu Bo. 2007.
\newblock \href {https://doi.org/10.1109/TPAMI.2007.1078} {A normalized levenshtein distance metric}.
\newblock \emph{IEEE Transactions on Pattern Analysis and Machine Intelligence}, 29(6):1091--1095.

\bibitem[{Zhang* et~al.(2020)Zhang*, Kishore*, Wu*, Weinberger, and Artzi}]{Zhang*2020BERTScore:}
Tianyi Zhang*, Varsha Kishore*, Felix Wu*, Kilian~Q. Weinberger, and Yoav Artzi. 2020.
\newblock \href {https://openreview.net/forum?id=SkeHuCVFDr} {Bertscore: Evaluating text generation with bert}.
\newblock In \emph{International Conference on Learning Representations}.

\bibitem[{Zhao et~al.(2023)Zhao, Zhou, Li, Tang, Wang, Hou, Min, Zhang, Zhang, Dong, Du, Yang, Chen, Chen, Jiang, Ren, Li, Tang, Liu, Liu, Nie, and rong Wen}]{Zhao2023ASO}
Wayne~Xin Zhao, Kun Zhou, Junyi Li, Tianyi Tang, Xiaolei Wang, Yupeng Hou, Yingqian Min, Beichen Zhang, Junjie Zhang, Zican Dong, Yifan Du, Chen Yang, Yushuo Chen, Z.~Chen, Jinhao Jiang, Ruiyang Ren, Yifan Li, Xinyu Tang, Zikang Liu, Peiyu Liu, Jianyun Nie, and Ji~rong Wen. 2023.
\newblock \href {https://api.semanticscholar.org/CorpusID:257900969} {A survey of large language models}.
\newblock \emph{ArXiv}, abs/2303.18223.

\end{thebibliography}

\clearpage

\appendix

\section{Appendix}
\label{sec:appendix}

\begin{figure*}[h!b]
\centering
\includegraphics[width=160mm]{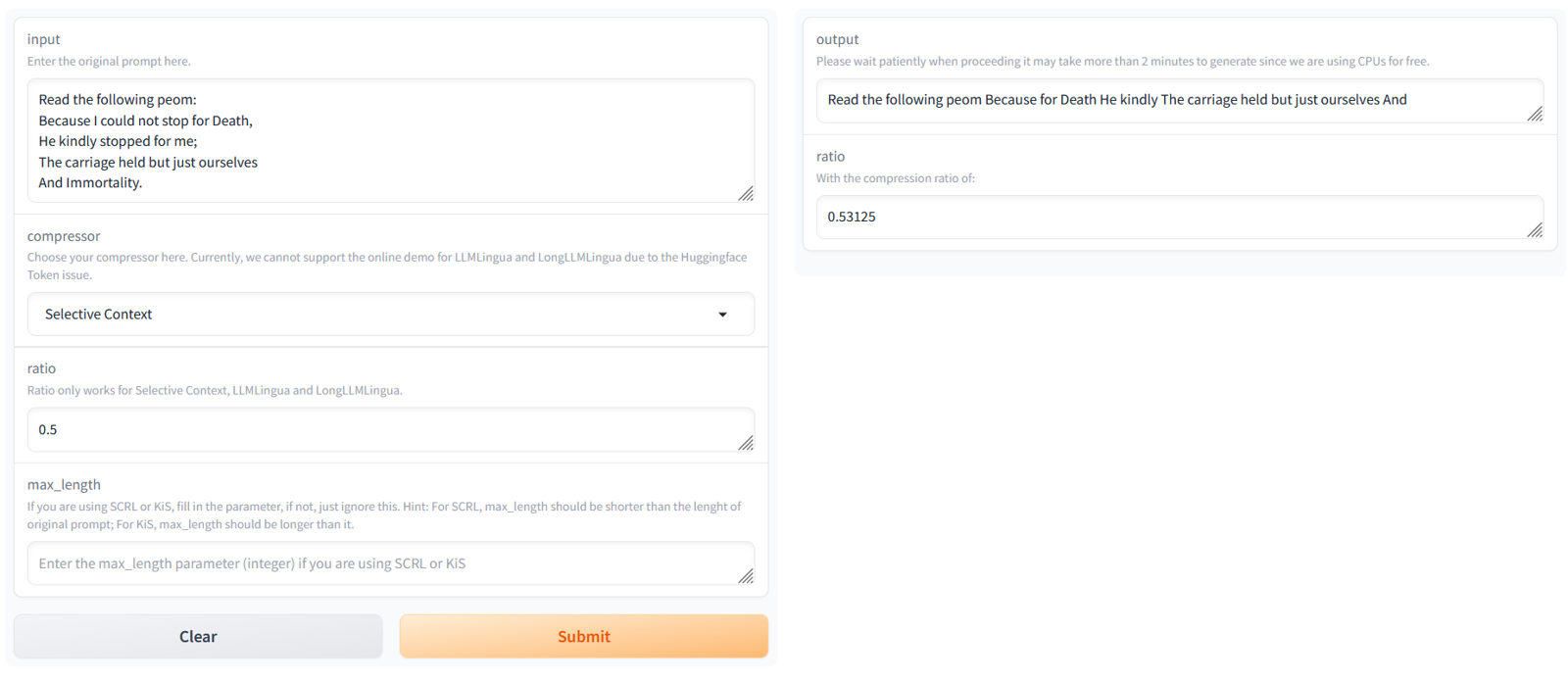}
\caption{Demonstration website.}
\label{fig2:env}
\end{figure*}

\subsection{Online Demonstration}

PCToolkit online demonstration is available on \url{https://huggingface.co/spaces/JerryLiJinyi/Prompt-Compression-Toolbox}. The guidance for online demonstration can be found in Appendix A.2.

\subsection{Guidance for Online Demonstration}

As shown in Figure~\ref{fig2:env}, follow the steps below to try our online demonstration.

\textbf{Step 1.} Enter the original prompt.

\textbf{Step 2.} Choose a compressor. Due to the Huggingface Token issue, we cannot provide online demonstrations for LLMLingua and LongLLMLingua compressors since they are based on LLaMA 2, for which a Huggingface Token is needed.

\textbf{Step 3.} Enter the compression ratio. As mentioned in \textbf{section 4}, the compression ratio is the proportion of context to be deleted. Compression ratio only works for Selective Context, LLMLingua and LongLLMLingua.

\textbf{Step 4.} If SCRL or KiS is chosen, \verb|max_length| parameter is needed to be specified manually. Precisely, for SCRL, \verb|max_length| represents the length of context window, so it should be less than the length of original context. As for KiS, \verb|max_length| represents the maximum length of input context. So, for KiS, \verb|max_length| should be longer than the original context.

\subsection{Datasets Tested on Different Compressors}

As shown in Table~\ref{ticks}, we evaluated all compressors on different datasets supported by PCToolkit.

\begin{table}[ht]
    \centering
    \begin{tabular}{p{1cm}m{1cm}<{\centering}m{1cm}<{\centering}m{1cm}<{\centering}m{0.8cm}<{\centering}m{0.8cm}<{\centering}}
    \toprule
        ~ & Selective Context & LLM-Lingua & Long-LLM-Lingua & SCRL & KiS \\ \hline
        GSM8K & \checkmark & \checkmark & \checkmark & \checkmark & \checkmark \\ \hline
        BBH & \checkmark & \checkmark & \checkmark & \checkmark & \checkmark \\ \hline
        BBC News & \multirow{2}{*}{\checkmark} & \multirow{2}{*}{\checkmark} & \multirow{2}{*}{\checkmark} & \multirow{2}{*}{\checkmark} & \multirow{2}{*}{\checkmark} \\ \hline
        Arxiv & \checkmark & \checkmark & \checkmark & \checkmark & \checkmark \\ \hline
        ShareGPT & \checkmark & \checkmark & \checkmark & \checkmark & \checkmark \\ \hline
        Gigaword & \checkmark & \checkmark & \checkmark & \checkmark & \checkmark \\ \hline
        DUC2004 & \checkmark & \checkmark & \checkmark & \checkmark & \checkmark \\ \hline
        BNC & \checkmark & \checkmark & \checkmark & \checkmark & \checkmark \\ \hline
        Broadcast & \checkmark & \checkmark & \checkmark & \checkmark & \checkmark \\ \hline
        Google & \checkmark & \checkmark & \checkmark & \checkmark & \checkmark \\
        \hline
        Long-Bench & ~ & \multirow{2}{*}{\checkmark} & \multirow{2}{*}{\checkmark} & ~ & ~ \\
        \bottomrule
    \end{tabular}
\caption{\label{ticks}
Datasets tested on different compressors.}
\end{table}

\end{document}